\documentclass{article}

\usepackage{PRIMEarxiv}

\usepackage[utf8]{inputenc} 
\usepackage[T1]{fontenc}    
\usepackage{hyperref}       
\usepackage{url}            
\usepackage{booktabs}       
\usepackage{amsfonts}       
\usepackage{nicefrac}       
\usepackage{microtype}      
\usepackage{lipsum}
\usepackage{fancyhdr}       
\usepackage{graphicx}       
\graphicspath{{media/}}     
\usepackage{array}
\usepackage{tcolorbox}
\newtcolorbox{textbox}[1]{colback=gray!5!white,colframe=black!45!white,fonttitle=\bfseries,title=#1}
\usepackage{multirow}

\title{Beyond Metrics: A Critical Analysis of the Variability in Large Language Model Evaluation Frameworks
}

\author{
  Marco AF Pimentel, Cl\'ement Christophe, Tathagata Raha, Prateek Munjal, \\
  \textbf{Praveen K Kanithi, Shadab Khan} \\
  \\
  M42 \\
  Abu Dhabi, UAE \\
  \texttt{\{mpimentel, cchristophe, traha, pmunjal, pkanithi, skhan\}@m42.ae} \\
}

\begin{document}
\maketitle

\begin{abstract}
As large language models (LLMs) continue to evolve, the need for robust and standardized evaluation benchmarks becomes paramount. Evaluating the performance of these models is a complex challenge that requires careful consideration of various linguistic tasks, model architectures, and benchmarking methodologies. In recent years, various frameworks have emerged as noteworthy contributions to the field, offering comprehensive evaluation tests and benchmarks for assessing the capabilities of LLMs across diverse domains. This paper provides an exploration and critical analysis of some of these evaluation methodologies, shedding light on their strengths, limitations, and impact on advancing the state-of-the-art in natural language processing.
\end{abstract}


\begin{figure}[!b]
  \centering
  \footnotesize
  \includegraphics[width=0.9\linewidth]{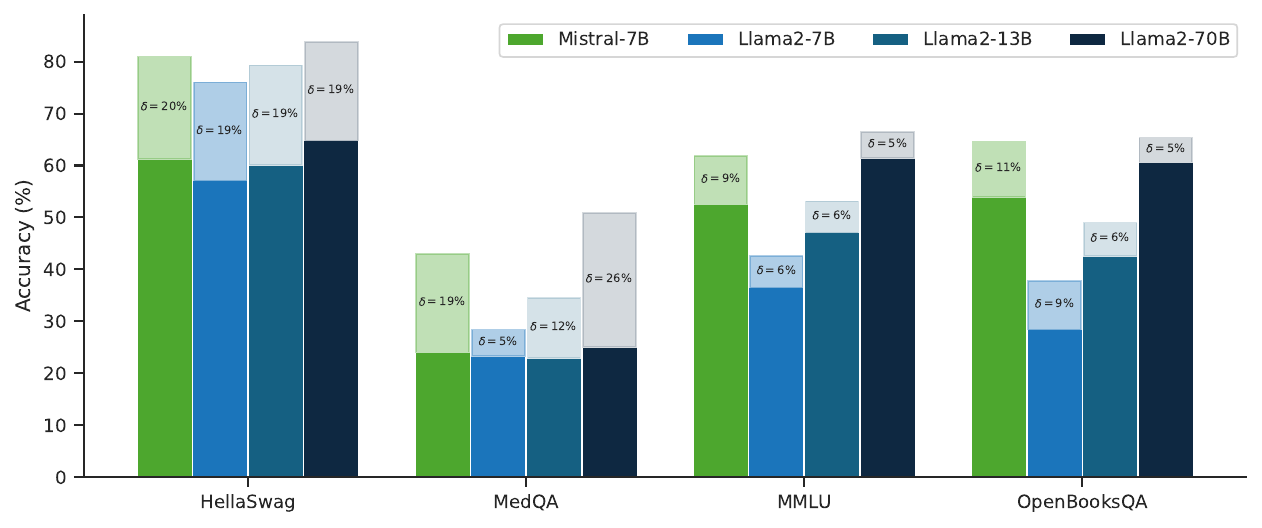}
  \caption{Performance disparities ($\delta$) of Llama2 (7B, 13B and 70B) and Mistral-7B on various benchmark datasets (in 0-shot setting). $\delta$ values are shown using lighter colors and they represent the variation observed in the accuracy metric across the benchmarks due to different evaluation frameworks.}
  \label{fig:fig1}
\end{figure}

\section{Introduction}
\label{sec:introduction}
The rapid advancements in the field of natural language processing (NLP) have been fueled by the development of increasingly sophisticated Large Language Models (LLMs). From OpenAI's GPT series \cite{openai2023gpt4} and Google's BARD \cite{chowdhery2022palm,driess2023palme} to a plethora of open-source models such as Llama \cite{touvron2023llama, touvron2023llama2}, the Falcon series \cite{falcon}, Jais \cite{sengupta2023jais} and ohers \cite{biderman2023pythia,dey2023cerebrasgpt,t52020}, these models showcase remarkable language understanding and generation capabilities. As the abilities of LLMs continue to expand, so does the demand for robust evaluation frameworks that can assess their linguistic aptitude and generalization capabilities \cite{chia2023instructeval, maynez2023benchmarking, gehrmann2022repairing}. A thorough evaluation is essential not only to benchmark the progress in model development but also to identify and address potential biases, handle ethical concerns, and ensure the responsible deployment of these models in real-world applications \cite{brown2020language, zhao2023survey, zhu2023large, laskar2023systematic, shevlane2023model}.

However, evaluating LLMs poses a unique set of challenges due to their parameter space, diverse applications, and sensitivity to the choice of evaluation tasks. Traditional metrics often fall short in capturing the nuances of language understanding, fluency, and context coherence exhibited by these models \cite{gehrmann2022repairing, liang2023holistic}. In addition, model performance is often affected by minor implementation details. It is exceedingly challenging to anticipate that results obtained from one codebase will seamlessly transfer to another \cite{Liu2023}. Frequently, research papers fail to provide the necessary code and/or sufficient details to replicate their evaluations fully. In response to this, the research community has witnessed the emergence of multiple evaluation frameworks for tracking the progress on specific NLP tasks across different benchmarks and to encourage multi-task learning with diverse datasets. To compare the performance of different LLMs, various leaderboards have also been created to rank LLMs according to their performance metrics (or scores) on existing or new evaluation benchmarks. For example, the GLUE \cite{wang-etal-2018-glue} and SuperGlue \cite{sarlin20superglue} benchmarks are widely used by language model developers. In addition, BIG-bench \cite{srivastava2023imitation}, HELM \cite{liang2023holistic}, EleutherAI's language model evaluation harness \cite{evalharness} and OpenCompass \cite{opencompass2023} frameworks have been introduced to study the capabilities of language models. 

As the demand for rigorous evaluation benchmarks intensifies, it becomes increasingly evident that the metrics employed are pivotal components. While these benchmarks provide invaluable insights into the capabilities of language models across diverse tasks, it is important to delve into the intricacies of metric calculation (or computation) to discern the subtleties of model behaviour. The significance of this aspect arises from the fact that metrics, often treated as objective measures, are subject to specific assumptions and methodologies during their calculation. These assumptions can significantly influence the interpretation of model performance and, consequently, the reliability of benchmark results. As the field progresses, the need for transparent and standardized reporting of metric calculation procedures becomes paramount to foster reproducibility and comparability across different studies.

This paper aims to delve into the understanding of different metric calculation methodologies employed by some of the most popular frameworks, providing some details on how these nuances can affect the performance of LLMs and the interpretability of evaluation results. Specifically, we describe an investigation into the accuracy metrics employed by prominent evaluation frameworks, detailing the methodology for their calculation. Our focus extends to evaluating four widely recognized open-source language models across question-answering datasets, which are comprised of those structured as multiple-choice scenarios, with the aim of providing a better understanding of their performance across these tasks.


\section{Background}
\label{sec:background}
In this investigation, we focus on multiple-choice question tests, in which for each question only one of the provided choices is the correct answer (see Box 1). This represents a rather simple task, in contrast to open-ended questions, for example. However, our examination reveals that even within this seemingly simple problem formulation, significant variability exists, stemming from subtle implementation details and disparities. Notably, the majority of these disparities arise from differences in the methods used to select the final answer option from the model's output. To help us formulate the problem, we focus on evaluating multiple-choice tasks using autoregressive language models, such as GPT (Generative Pre-trained Transformer) like architectures, which underpins the architecture of the models mentioned above. 

\begin{textbox}{Box 1: Example multiple-choice question with answer.}
One of the reasons that the government discourages and regulates monopolies is that 
~

~

Choices: 

A. producer surplus is lost and consumer surplus is gained. 

B. monopoly prices ensure productive efficiency but cost society allocative efficiency. 

C. monopoly firms do not engage in significant research and development. 

D. consumer surplus is lost with higher prices and lower levels of output.
~

~

Correct answer: D
\end{textbox}

We frame mathematically the evaluation of multiple-choice questions (MCQs) using LLMs as follows. Let \(Q\) be the set of multiple-choice questions, each denoted as \(q_i\), where \(i\) ranges from 1 to \(N\), where \(N\) represents the total number of questions. For each question \(q_i\), there exists a set of answer choices \(A_i = \{a_{i1}, a_{i2}, ..., a_{ik}\}\), where \(k\) is the number of choices for question \(q_i\), with \(2\leq k \leq5\). The goal is to assess the model's performance in selecting the correct answer from the provided choices. Let \(c_i\) be the correct answer for question \(q_i\), where \(c_i \in A_i\) . The model's predicted answer for \(q_i\) is denoted as \(\hat{c}_i\). We note that ideally \(\hat{c}_i \in A_i\), but it may not always be the case; i.e., the model may generate an answer that is not necessarily part of the set of choices \(A_i\). From this point onward, to simplify further our notation, we refer to a single question of a dataset as \(q\) and set of answer choices for that question as \(A = \{a^{1}, a^{2}, ..., a^{k}\}\); i.e., we ommit the index \(i\).

LLMs operate by accepting a textual input, i.e., a question/instruction (also called a 'prompt') which is segmented into tokens (typically, sub-words). Let the input prompt be represented as \(q_{0:m}\), where \(m+1\) is the number of tokens of the input prompt. From this tokenized input, the language model generates a conditional probability distribution \(P(q_{m+1}|q_{0:m})\) over the vocabulary of tokens of the model for predicting the next token. This allows LLMs to estimate the likelihood of any token as continuation of the input prompt. By appending the selected token to the prompt and iterating this process, the model generates subsequent tokens, enabling the creation of entire sentences as continuations of the initial input prompt. Let \(q_{m+1:n_k}\) be the \(k\)-th possible continuation sequence generated by the LLM, with \(n_k-m\) tokens.  

Hence, for model evaluation, two primary approaches emerge:

\begin{itemize}
    \item Token probability comparison: this involves obtaining the probabilities \(P(q_{m+1}|q_{0:m})\) for given sets of tokens and comparing these probabilities for the predefined choices (\(A\)); i.e., the evaluation metric can be constructed by assessing the relative likelihoods of various token groups as continuations of the prompt.
    \item Text generation comparison: one can obtain a text generation from the model using the iterative process described above and then compare the generated text to the various predefined possible choices; this provides a holistic assessment of the model's ability to generate coherent and contextually relevant continuations.
\end{itemize}

Both token-level assessments and holistic text generation analyses underscore popular evaluation frameworks. In the following sections, we detail three of these frameworks and their implementations: OpenCompass, EleutherAI's language model evaluation harness, and Stanford University's HELM. 

\subsection{OpenCompass}
OpenCompass, an open-source evaluation framework for language models\footnote{OpenCompass' evaluation framework also supports multimodal models, which is not within the scope of this investigation.}, uses the token probability comparison approach for extracting the prediction of the model \cite{opencompass2023}. Specifically, the probabilities predicted by the model for all possible answers are compared, such that the probability for the option \(a^k\) is given by $P(q_{m+1} | q_{0:m})$; in the example above $a^k \in \{``A", ``B", ``C", ``D"\}$, i.e., the first letter corresponding to each choice. We note that the framework uses perplexity as the key metric, rather than relying solely on (log) likelihood. 

In order to reduce the likelihood of the model generating responses that fall outside the intended range of answers, a ``few shots'' approach is typically used, in which the prompt is augmented with one or more instances/examples with their correct answers as well (see Appendix B). In instances where the model might have otherwise generated an unrelated word (or token), the inclusion of a few shots attempts to ensure that the model is guided by known examples and by better understanding of the expected behaviour. The approach of incorporating a handful of examples in the prompt typically improves the model's performance and is a standard evaluation method, as demonstrated in assessments such as MMLU, where five shots (i.e., five examples) are prepended to each prompt for evaluation across various benchmarks \cite{hendryckstest2021}.

\subsection{LM Evaluation Harness}
\label{ssec:harness}
The evaluation harness framework from EleutherAI \cite{evalharness} also uses the token probability comparison methodology. In this case, however, it computes the likelihood of the entire continuation sequence using the process described above (i.e., it uses the full answer sequence which contains the letter followed by the text of the answer). In its simplest method, for each \(k\) choice in \(A\), the likelihood is determined using 
$$\sum_{j=m+1} ^{n_k} \log P(q_j | q_{0:j-1}),$$ 
where the aggregation of the probabilities is achieved by summing the log of the individual probabilities for numerical stability\footnote{This approach is used in \cite{evalharness} in all multiple choice tasks and tagged as ``acc".}. Conceptually, this entails determining the probability of a generated sequence, sampled from the given prompt, incorporating the specific continuation (or choice) under consideration (in this case, of the entire answer). The few-shot prompt is generally similar to the approach described in the previous section. Although straightforward, this method does not take into account (possible) differences in length between the predefined choices that may be substantical (i.e., $n_k$ can vary substantially for each $k$). This can bias the model toward favoring longer choices as they tend to have higher log probabilities (or likelihoods). 

In order to tackle this problem, the framework's authors introduced a normalization step in which the overall likelihood is divided by a measure of the length of the answer sequence. This can be accomplished in two ways: either using token-length normalization or byte-length normalization. In the former, the normalized likelihood of the \(k\)-th option is determined using the average log probability per token: 
$$\sum_{j=m+1} ^{n_k} \log P(q_j | q_{0:j-1}) / (n_k-m).$$ 
However, the authors noted that this approach is not tokenization agnostic, which means that two models with distinct tokenization procedures (and/or vocabulary sizes), despite assigning the same log likelihood to every single input string, may yield different token-length normalized log likelihoods. 

The byte-length normalization approach attempts to normalize the likelihood by computing the average log probability per character, which ensures that it is tokenization agnostic\footnote{This approach is also used in \cite{evalharness} in all multiple choice tasks and is tagged as ``acc\_norm".}. In this case, the normalized likelihood of the \(k\)-th option is determined using 
$$\sum_{j=m+1} ^{n_k} \log P(q_j | q_{0:j-1}) / \sum_{j=m+1} ^{n_k} L_{q_j},$$
where $L_{q_j}$ is the number of bytes represented by the token $q_j$. In practice, rather than using the number of bytes, the number of characters in $a^k$ is used for normalization. 

\subsection{HELM}
Another popular framework is the Holistic Evaluation of Language Models (or HELM) project \cite{liang2023holistic}. The method used for evaluating the model using MCQs in HELM is different from the implementations described in the previous two approaches. Instead of comparing the token probabilities from the given answer choices, HELM utilizes the model's next token output probabilities to generate text. This generated text is then compared to the expected answer. Specifically for MCQ tasks, HELM implements metric functions such as exact match to assess the correspondence between the generated text and the correct answer. This approach allows for a more natural evaluation of the model's understanding and ability to generate relevant responses, rather than simply selecting the most probable option from a predefined set of choices.

While the evaluation methodology takes a distinct approach, the few-shot prompt remains generally similar. However, for a given instance, we note that if the model assigns the highest probability to a token that deviates from the intended range of answers, even though it is not part of the set of choices, the model's response would be deemed incorrect, resulting in a lack of scoring for that particular instance. In other words, the evaluation process hinges on the model's ability to prioritize the correct tokens within the specified answer choices. In the event that the model, despite a generally similar few-shot prompt, allocates the highest probability to a token outside the intended range, the response in deemed incorrect. This is substantially different from the methods described above, in which only the probabilities associated with the given set of answers are included for computing the model's performance.

We also note that frameworks such as HELM \cite{liang2023holistic} and Langtest \cite{langtest} attempt to enhance the evaluation process of LLMs by offering a broader array of tasks and metrics (in addition to accuracy). These frameworks offer a comprehensive set of evaluation criteria that delves into various aspects of language understanding and generation, providing additional metrics such as calibration, robustness, fairness, bias, toxicity, and efficiency. Such metrics are out of the scope of this study.

\section{Methods}
\label{sec:methods}
In this study, we focus on two prominent evaluation frameworks, OpenCompass and Eval harness. As described above, these frameworks utilize a token-probability comparison method to assess LLMs in the context of multiple-choice question answering benchmarks. Our investigation centers on providing a detailed account of the accuracy metrics obtained through these frameworks, with particular attention to four widely recognized LLMs, for quantifying the performance of such models. We analyse the results across four popular benchmarks, with the goal of offering a better understanding of the performance and variability in performance of the selected models within the defined evaluation paradigms.

\subsection{Evaluation tasks and datasets}
In this section, we describe the test datasets used to evaluate the models on different tasks. When selecting these datasets, preference was given to widely acknowledged sources used across various domains, such as general and medical contexts. Additionally, these encompass a range of context lengths and types of reasoning. Table \ref{tab:datasets-count} includes a summary of the datasets used in this study.

\begin{table}[t]
\footnotesize
\caption{Question-answering evaluation datasets.}
\centering
\begin{tabular}{lm{2cm}m{8cm}}
\toprule
\rule{0pt}{9pt}
Dataset name & Count & Dataset description \\
\midrule
\rule{0pt}{9pt}
HellaSwag    & 10,042 & Common-sense reasoning about the physical world \\
\rule{0pt}{9pt}
MedQA        & 1,273  & General medical knowledge from medical board exams \\
\rule{0pt}{9pt}
MMLU         & 14,042 & Knowledge and problem-solving tasks covering various topics \\
\rule{0pt}{9pt}
OpenBookQA   & 500 & Multi-step, common-sense reasoning and text comprehension \\
\bottomrule
\end{tabular}
\label{tab:datasets-count}
\end{table}

\textbf{HellaSwag.} This dataset was introduced in 2019 to test commonsense natural language inference about physical situations \cite{zellers2019hellaswag}. HellaSwag\footnote{"HellaSwag" is short for Harder Endings, Longer contexts, and Low-shot Activities for Situations With Adversarial Generations.} employed "Adversarial Filtering", in which the idea is to (machine) generate challenging incorrect answers for a multi-choice test setting (see Box A1). These incorrect answers are dubbed 'adversarial endings'. Though humans score above 95.6\% on HellaSwag, initial state-of-the-art models struggled (with accuracies lower than 48\% back in 2019).

\textbf{MedQA.} We included a domain-specific dataset, MedQA, originally designed for addressing medical problems \cite{jin2021disease}. The dataset comprises free-form multiple-choice question-answers which were sourced from professional medical board exams (Box A2). 

\textbf{MMLU.} The Measuring Multitask Language Understanding (MMLU) benchmark \cite{hendryckstest2021} aimed to introduce a comprehensive assessment of LLMs across 57 subjects, including elementary mathematics, US history, computer science, law, and others (Box A3).

\textbf{OpenBookQA.} Mihaylov et al. (2018) \cite{OpenBookQA2018} presented a question-answering dataset that attempts to probe a deeper understanding of a topic (Box A4). Modeled after open book exams for assessing human understanding of a subject, it contains questions that require multi-step reasoning, use of additional common and commonsense knowledge, and rich text comprehension.

\subsection{Language model architectures}
We employed two distinct family of language models, namely Llama2 \cite{touvron2023llama2} and Mistral \cite{jiang2023mistral}, to evaluate their performance within the context of the OpenCompass and Eval harness benchmarks. These pretrained generative text models have been shown to perform well across various benchmarks. 

Llama2 models, part of the family of language models developed by Meta AI, are a set of LLMs with varying size, ranging from 7 billion to 70 billion parameters. It is an auto-regressive language model based on the transformer decoder architecture with some notable differences from models like GPT-3. For example, Llama2 employs the SwiGLU activation function rather than ReLU and uses rotary positional embeddings in lieu of absolute learnable positional embeddings \cite{touvron2023llama2}. The latest release of Llama2 also introduces architectural refinements geared towards enhanced performance, extending the context length to up to 4,096 tokens. Bigger models (70B) use Grouped-Query Attention (GQA) to better leverage long sequences and improved inference scalability. 

Mistral-7B (v0.1), introduced by Mistral AI, is a 7.3-billion parameter model with a similar architecture to that of Llama2 \cite{jiang2023mistral}. It also uses grouped-query attention which enhances the inference process by caching key and value vectors for previously decoded tokens in the sequence, thereby reducing processing time. In addition, it uses a sliding window-based attention mechanism which replaces full attention, characterized by square compute cost. In this mechanism, each token can attend to at most 4,096 tokens from the preceding layer, resulting in a linear compute cost. This implementation enhances Mistral-7B's capability to handle long sequences (upto 32k), allowing higher layers to access historical information beyond the 4,096-token context window size.

In this investigation, from the family of Llama2 models we use the 7-billion (Llama2-7B), 13-billion (Llama2-13B) and 70-billion-parameter models (Llama2-70B) along with Mistral-7B.

\subsection{Evaluation metrics}
The assessment of the three models on the aforementioned datasets involves the utilization of the accuracy metrics derived from OpenCompass and Eval harness. The final accuracy for a dataset is determined by the percentage of questions answered correctly. To determine this accuracy, we can follow different methods to identify the correct option selected by a model (as described in the previous section). Our focus centers on the accuracy metrics computed using the different methods employed by different evaluation frameworks. We attend to the following methods for this purpose:

\begin{itemize}
\item OpenCompass' accuracy (denoted OC accuracy): this approach invoves extracting the model's prediction by assessing the next token probabilities and determining whether the selected choice (that with the highest likelihood) is the correct answer (as described above and used in \cite{opencompass2023});

\item Raw (unnormalized) accuracy (Raw accuracy): similarly to the previous method, it involves comparing token probabilities, but in this case, the probabilities of the full answers' sequences are used to determine the correct option, using the sum of the log likelihoods of all tokens; this corresponds to the method used and reported in the eval harness' framework (see section \ref{ssec:harness});

\item Token-normalized accuracy (T-norm accuracy): this method involves normalizing the likelihhood for each answer (as obtained for the metric above) by dividing the sum by the number of tokens to avoid giving too much "weight" to longer answers (section \ref{ssec:harness});

\item Byte-normalized accuracy (B-norm accuracy): this method also attempts to avoid biasing the model toward favoring longer choices by normalizing the likelihood using the average log probability per character (using the number of characters in the full answer sequence); as mentioned in section \ref{ssec:harness}, it is also used and reported in the eval harness' framework.

\end{itemize}

The evaluation is carried out in a zero-shot setting, meaning that no examples are added to the prompt. Also, to ensure consistency and mitigate the impact of variations in prompts on model results, a standardized prompt design was employed across the evaluation frameworks and methods. Refer to Appendix B for further details on the adopted prompt design for each dataset.

\section{Results}
\label{sec:results}

The performance of the models across selected benchmark datasets is summarized in the Table \ref{tab:main-results}, highlighting the accuracy metrics obtained within each one of the four evaluation methods.

\begin{table}[t]
\footnotesize
\caption{0-shot performance comparison of the different models on the selected benchmark datasets. Accuracy results (in \%) are shown according to each of the four evaluation methods used for determining the correct answer.}
\centering
\begin{tabular}{@{}llcccc@{}}
\toprule
                                    &            & HellaSwag       & MedQA         & MMLU          & OpenBookQA     \\ \midrule
\multirow{4}{*}{OC accuracy}        & Mistral-7B & 78.9 & 24.0 & 61.8 & \textbf{64.6}  \\
                                    & Llama2-7B  & 74.0            & 23.3          & 42.6          & 32.0           \\ 
                                    & Llama2-13B & 77.5            & 22.9          & 53.1          & 42.6           \\
                                    & Llama2-70B & \textbf{82.3}                & \textbf{25.1} & \textbf{66.5} & 61.2              \\ \midrule
\multirow{4}{*}{Raw accuracy}       & Mistral-7B & 61.2            & 42.9          & 58.2          & 54.0           \\
                                    & Llama2-7B  & 57.1            & 28.4          & 38.2          & 28.4           \\
                                    & Llama2-13B & 60.1            & 34.5          & 51.0          & 47.5           \\
                                    & Llama2-70B & \textbf{64.8}   & \textbf{50.8} & \textbf{65.8} & \textbf{65.4}  \\ \midrule
\multirow{4}{*}{T-norm accuracy}    & Mistral-7B & 80.0            & 39.8          & 54.1           & 56.2           \\
                                    & Llama2-7B  & 74.9            & 23.8          & 37.7          & 37.6           \\
                                    & Llama2-13B & 78.3            & 30.8          & 48.3          & 47.8           \\
                                    & Llama2-70B & \textbf{82.8}   & \textbf{47.1} & \textbf{62.8} & \textbf{63.6}  \\ \midrule
\multirow{4}{*}{B-norm accuracy}    & Mistral-7B & 81.0            & 38.3          & 52.5          & 57.0           \\
                                    & Llama2-7B  & 76.0            & 25.4          & 36.5          & 37.8           \\
                                    & Llama2-13B & 79.5            & 30.6          & 47.1          & 49.0           \\
                                    & Llama2-70B & \textbf{83.8}   & \textbf{47.0} & \textbf{61.4} & \textbf{60.6}  \\  
\bottomrule
\end{tabular}
\label{tab:main-results}
\end{table}

For each evaluation scenario, we note that Llama2-70B consistently outperforms other Llama2 (smaller) variants and Mistral-7B across all benchmark datasets. However, within each benchmark dataset, a substantial variability in the performance of the different models across the four methods is observed (Figure \ref{fig:fig1}). As examples, for MMLU, the performance of Mistral-7B ranges from 61.4\% and 65.8\%; while for HellaSwag, the performance of Llama2-70B fluctuates between 64.8\% and 83.8\%.

We also note that the effect of the normalization methods (e.g., B-norm accuracy) is not consistent across the benchmark datasets; while for HellaSwag, the normalization-based accuracy metrics are higher than the raw-based accuracy metric, which does not take into account the normalization of the log likelihoods of the responses (according to their length), the opposite behaviour is observed for other datasets (Table \ref{tab:main-results}). To delve deeper into the factors influencing correct option selection, we investigate the impact of normalizing response likelihoods. For each question-answer pair, we assessed the length of options and compared the length of the correct option with that of the wrong options (this is set to be the median length of the wrong options). 

\begin{figure}[!t]
  \centering
  \footnotesize
  \includegraphics[width=0.9\linewidth]{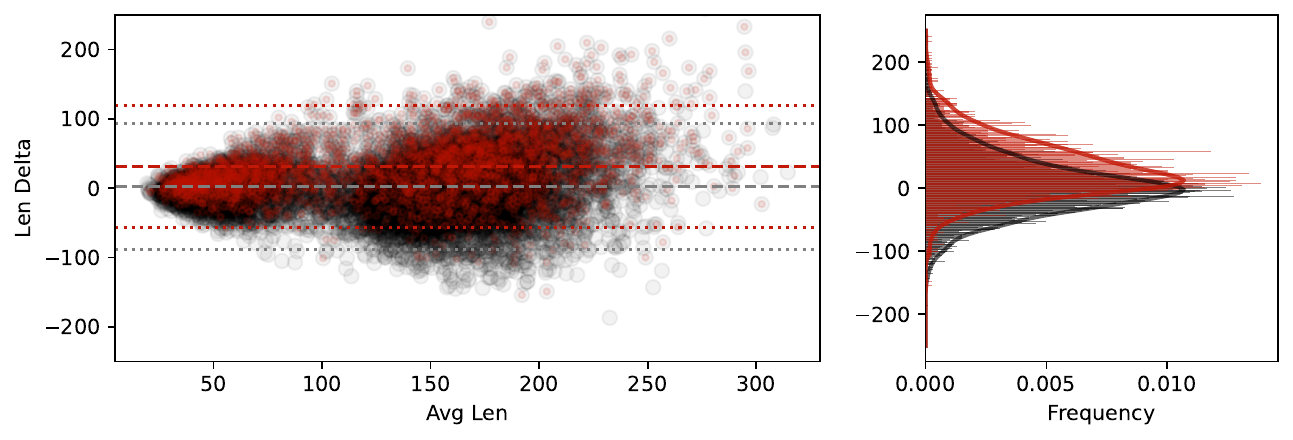}
  \includegraphics[width=0.9\linewidth]{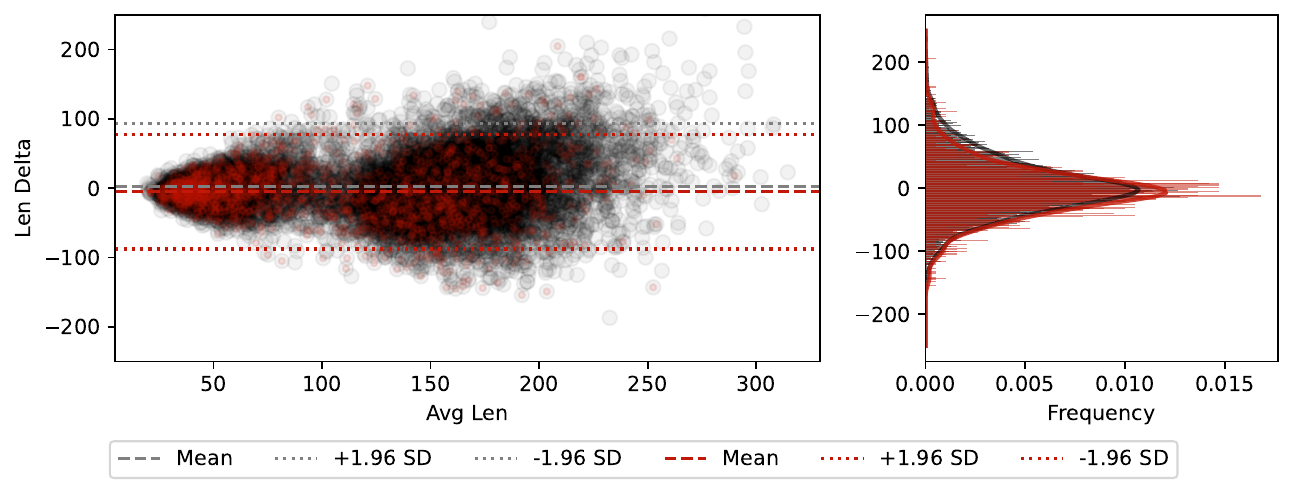}
  \caption{Bland-Altman plots (left) and frequency plots or histograms (left) of the length difference between correct and wrong options for the \textbf{HellaSwag} benchmark dataset. The length difference for the entire dataset is shown in black (in both top and bottom panels). In the top panel, the length differences for the instances in which Mistral-7B incorrectly selected the wrong option in the unnormalized likelihood method (raw-based accuracy) are overlayed in red. On the bottom panel, the length differences for the instances in which Mistral-7B incorrectly selected the wrong option in the B-norm accuracy method are overlayed in red.}
  \label{fig:fig2}
\end{figure}

The Bland-Altman plot (depicted in black in Figure \ref{fig:fig2}) illustrates the length difference between right and wrong options for the (whole) HellaSwag dataset. No significant inherent bias in the length of the correct options compared to the length of the wrong options is observed (and the same is observed for the other datasets too; figure not included in the manuscript). I.e., the lengths of correct and wrong answer options of the records included in the benchmark dataset are similar. Notably, the figure also shows the length difference (in red) for those instances in which Mistral-7B incorrectly selected the wrong option according to the unnormalized likelihood method (top panel) and B-norm likelihood method (bottom panel). 

\begin{figure}[!t]
  \centering
  \footnotesize
  \includegraphics[width=0.9\linewidth]{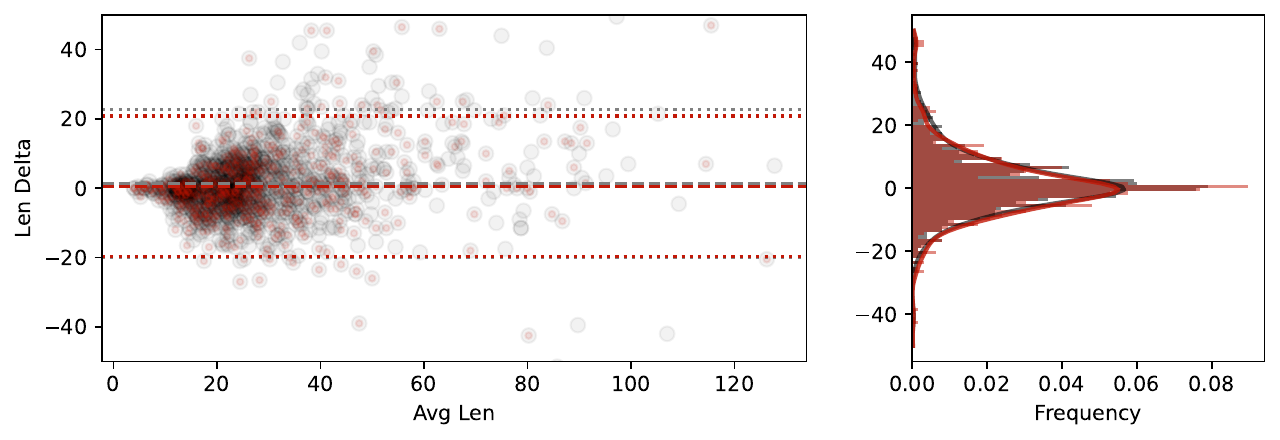}
  \includegraphics[width=0.9\linewidth]{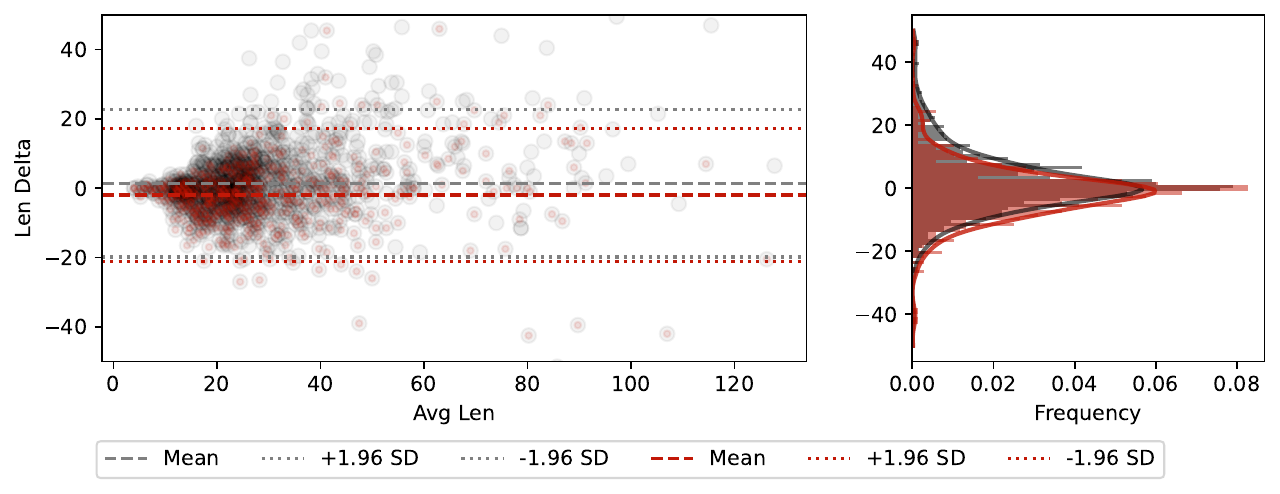}
  \caption{Bland-Altman plots (left) and frequency plots or histograms (left) of the length difference between correct and wrong options for the \textbf{MedQA} benchmark dataset. The length difference for the entire dataset is shown in black (in both top and bottom panels). In the top panel, the length differences for the instances in which Llama2-70B incorrectly selected the wrong option in the unnormalized likelihood scenario (raw-based accuracy) are overlayed in red. On the bottom panel, the length differences for the instances in which Llama2-70B incorrectly selected the wrong option in the B-norm accuracy scenario are overlayed in red.}
  \label{fig:fig3}
\end{figure}

Figure \ref{fig:fig3} shows the results of a similar analysis for the MedQA benchmark dataset.

\section{Discussion}
In this paper, we aimed to explore the details of metric calculation methodologies employed by prominent evaluation frameworks and their implications on the performance assessment of LLMs and the subsequent interpretability of evaluation results. Specifically, we scrutinized the accuracy metrics, describing the intricacies of their calculation methodologies. Extending our examination beyond theoretical considerations, we directed our efforts towards the evaluation of widely recognized open-access language models. This evaluation encompassed different question-answering datasets, including those structured as multiple-choice scenarios. 

The results of our analyses shed light on some critical aspects of the evaluated LLMs and the associated methodologies. Notably, considering each method individually, Llama2-70B consistently demostrates superior performance across all benchmark datasets when compared to both its smaller variants and Mistral-7B (Table \ref{tab:main-results}). This has been observed in other studies, in which the performance of Llama2-70B has been deemed to be superior in diverse evaluation benchmark tasks, including question-answering tasks \cite{touvron2023llama2, jiang2023mistral}.

Nevertheless, despite this overall superiority, the results from our analysis demonstrate a substantial variability within each benchmark dataset across the four evaluation methods (Figure \ref{fig:fig1}). The observed fluctuations in performance, ranging between 5\% and 26\%, highlight the sensitivity of the reported accuracy metrics to the method-specific implementations. 

In evaluation frameworks such as eval-harness by Eleuther AI \cite{evalharness}, for calculating accuracy of a model, the likelihood of the (complete) response (for each option) is determined, and the option with the highest likelihood is deemed to be the correct one. In order to avoid the introduction of a bias toward favoring shorter answers, eval-harness has introduced a normalization step, in which the overall likelihood for each choice is divided by a measure of the length of the choice's sequence. These different approaches, as expected, produce different results.  

In addition, Figure \ref{fig:fig2} highlights the impact of normalization in the selection bias of the correct option by a model (in red) on HellaSwag. In the top panel (without normalization of the response's likelihood), a substantial bias is evident, with the model preferring shorter answers; whereas, if normalization of the response's likelihood (bottom panel) is employed, the bias is noticeably reduced, underscoring the effect of normalization on enhancing the reliability of the response selection, and ultimately the model's accuracy. However, it is important to note that the impact of normalization on the bias is not universally consistent across all datasets. In contrast to the observed mitigation of bias in the HellaSwag dataset, the same analysis for other datasets reveals that the normalization process seems to introduce a bias rather than reducing it in the selection of the correct option. This can be observed, for example, in the MedQA dataset (Figure \ref{fig:fig3}). In this case, the normalization of response likelihoods appears to contribute to a discernible bias in the selection of the correct option.

The impact of the responses' likelihood normalization methods on the accuracy metrics reveals a lack of consistent behaviour across benchmark datasets. While for some datasets, the normalization step appears to reduce the bias in the choice selection and produce better performance results, for other the normalization step seems to introduce a bias in the choice selection and, hence, provide less accurate results.

This variation in performance across frameworks highlights the importance of comprehensive benchmarking. The performance of an LLM is typically a function of its architecture, the model’s training data and the extent to which it has been fine-tuned to a particular domain of knowledge; i.e., one model may be exceedingly good at answering medical questions, but it barely exceeds “Hello World” in a Python programming test. Notwithstanding this, the results obtained in this investigation, as also observed in \cite{zhou2023dont}, suggest that the performance of an LLM is also highly dependent on the methodology used and implementation details employed to evaluate them (even when using the same benchmark dataset). 

To conclude, evaluation frameworks play an important role in the assessment and enhancement of LLMs. They are pivotal in addressing key challenges related to performance evaluation, offering valuable insights for model development, and enhancing transparency. By providing a multi-dimensional perspective on LLMs, existing frameworks make substantial contributions to the advancement and comprehension of these intricate AI systems. As the demand for rigorous evaluation of LLMs intensifies, it becomes imperative not only to scrutinize the metrics themselves but also to expose the methodologies of their calculation. The diversity in evaluation frameworks also introduces a considerable challenge, as each framework may employ distinct metrics and computation methodologies tailored to its specific objectives. Understanding the underlying assumptions and approaches during metric calculation is crucial for interpreting and comparing the results effectively. Variability in metric definitions across frameworks can introduce ambiguity, making it difficult to draw meaningful cross-model and cross-study comparisons. Additionally, certain metrics may inadvertently favor specific model characteristics or exhibit sensitivity to dataset idiosyncrasies. Consequently, in the absence of thorough evaluation frameworks, researchers must delve into the nuances of metric definitions and computation procedures to grasp the full context of reported results, ensuring an informed understanding of LLM performance. This shift towards a more transparent and methodologically explicit evaluation paradigm is pivotal for fostering reproducibility and advancing our collective understanding of the capabilities and limitations of large language models.

\section*{Acknowledgments}
This study has been supported by M42.

\bibliographystyle{unsrt}  
\bibliography{references}  

\newpage

\section*{Appendix A: Datasets' examples}

An example instance from each one of the four datasets used in this study is provided below.

\begin{textbox}{Box A1: Example test record from the HellaSwag dataset.}
A bearded man is seen speaking to the camera and making several faces. The man 
~

~

\textbf{Correct answer:} \\
- then holds up a razor and begins shaving his face.
~

~

\textbf{Incorrect answers:} 

- then switches off and shows himself via the washer and dryer rolling down a towel and scrubbing the floor. 

- then rubs and wipes down an individual's face and leads into another man playing another person's flute.

- is then seen eating food on a ladder while still speaking.
\end{textbox}

\begin{textbox}{Box A2: Example question/answer from the MedQA dataset (the option in bold is deemed to be the correct answer).}
A 23-year-old pregnant woman at 22 weeks gestation presents with burning upon urination. She states it started 1 day ago and has been worsening despite drinking more water and taking cranberry extract. She otherwise feels well and is followed by a doctor for her pregnancy. Her temperature is 97.7°F (36.5°C), blood pressure is 122/77 mmHg, pulse is 80/min, respirations are 19/min, and oxygen saturation is 98\% on room air. Physical exam is notable for an absence of costovertebral angle tenderness and a gravid uterus. Which of the following is the best treatment for this patient?
~

~

(A) Ampicillin.~~~~(B) Ceftriaxone~~~~(C) Ciprofloxacin~~~~(D) Doxycycline~~~~\textbf{(E) Nitrofurantoin} 
\end{textbox}

\begin{textbox}{Box A3: Example test from MMLU dataset's Microeconomics task (the option in bold is deemed to be the correct answer).}
One of the reasons that the government discourages and regulates monopolies is that 
~

~

(A) producer surplus is lost and consumer surplus is gained.

(B) monopoly prices ensure productive efficiency but cost society allocative efficiency. 

(C) monopoly firms do not engage in significant research and development. 

\textbf{(D) consumer surplus is lost with higher prices and lower levels of output.} 
\end{textbox}

\begin{textbox}{Box A4: Example question/answer from the OpenBookQA dataset (the option in bold is deemed to be the correct answer).}
Which of these would let the most heat travel through? 
~

~

(A) a new pair of jeans. 

\textbf{(B) a steel spoon in a cafeteria.} 

(C) a cotton candy at a store. 

(D) Doxycycline. 
\end{textbox}


\newpage

\newpage

\section*{Appendix B: Input prompts}
In order to guarantee similar inputs to the model using either benchmark framework, we standardized the prompt for each one of the datasets. We show example prompts for each dataset.

\begin{textbox}{Box B1: Example prompt input and outputs from MMLU dataset's Microeconomics task in Eval-harness.}
------------------------------------------------------------------------------------------------------------------------------------ 

One of the reasons that the government discourages and regulates monopolies is that 

(A) producer surplus is lost and consumer surplus is gained. 

(B) monopoly prices ensure productive efficiency but cost society allocative efficiency. 

(C) monopoly firms do not engage in significant research and development. 

(D) consumer surplus is lost with higher prices and lower levels of output. 

Answer: 

------------------------------------------------------------------------------------------------------------------------------------ 
\textbf{Possible answers:} 

(A) producer surplus is lost and consumer surplus is gained. 

(B) monopoly prices ensure productive efficiency but cost society allocative efficiency. 

(C) monopoly firms do not engage in significant research and development.

(D) consumer surplus is lost with higher prices and lower levels of output. 
\end{textbox}

\begin{textbox}{Box B2: Example prompt input and outputs from MMLU dataset's Microeconomics task in OpenCompass.}
------------------------------------------------------------------------------------------------------------------------------------ 

One of the reasons that the government discourages and regulates monopolies is that 

(A) producer surplus is lost and consumer surplus is gained. 

(B) monopoly prices ensure productive efficiency but cost society allocative efficiency. 

(C) monopoly firms do not engage in significant research and development. 

(D) consumer surplus is lost with higher prices and lower levels of output. 

Answer: 

------------------------------------------------------------------------------------------------------------------------------------ 

\textbf{Possible answers:} 

A. 

B. 

C. 

D. 
\end{textbox}

\end{document}